\documentclass[english,a4paper,11pt]{article}
\usepackage{algorithm}
\usepackage{algorithmic}
\usepackage{fullpage}
\usepackage{adjustbox}
\usepackage{lscape}

\usepackage{amsxtra, amsfonts, amssymb, amstext}
\usepackage{amsthm}
\usepackage{booktabs}
\usepackage{longtable}
\usepackage{nicefrac}
\usepackage{xspace}
\usepackage[noadjust]{cite}
\usepackage{url}
\usepackage{graphics,color}
\usepackage{hyperref}
\usepackage[algo2e,ruled,vlined,linesnumbered]{algorithm2e}
\newcommand{\assign}{\leftarrow}
\usepackage{graphicx}
\usepackage{wrapfig}
\usepackage{balance}

\newcommand{\R}{\mathbb{R}}

\newcommand{\om}{\textsc{OneMax}\xspace}
\newcommand{\onemax}{\om}

\newcommand{\lo}{\textsc{LeadingOnes}\xspace}
\newcommand{\leadingones}{\lo}

\newcommand{\opl}{$(1+\lambda)$~EA\xspace}

\DeclareMathOperator{\Bin}{Bin}
\DeclareMathOperator{\mut}{flip}

\def\BibTeX{{\rm B\kern-.05em{\sc i\kern-.025em b}\kern-.08em
    T\kern-.1667em\lower.7ex\hbox{E}\kern-.125emX}}
\begin{document}

\title{Interpolating Local and Global Search by Controlling the Variance of Standard Bit Mutation\thanks{This work was supported by the Chinese scholarship council (CSC No. 201706310143), the Paris Ile-de-France Region, COST Action CA15140, and by a public grant as part of the Investissement d'avenir project, reference ANR-11-LABX-0056-LMH, LabEx LMH, in a joint call with Gaspard Monge Program for optimization, operations research and their interactions with data sciences.}}

\author{
Furong Ye, LIACS, Leiden University, The Netherlands\\
Carola Doerr, LIP6, Sorbonne University, CNRS, Paris, France\\
Thomas B\"ack, LIACS, Leiden University, The Netherlands
}

\maketitle

\begin{abstract}
A key property underlying the success of evolutionary algorithms (EAs) is their global search behavior, which allows the algorithms to ``jump'' from a current state to other parts of the search space, thereby avoiding to get stuck in local optima. This property is obtained through a random choice of the radius at which offspring are sampled from previously evaluated solutions. It is well known that, thanks to this global search behavior, the probability that an EA using standard bit mutation finds a global optimum of an arbitrary function $f:\{0,1\}^n \to \mathbb{R}$ tends to one as the number of function evaluations grows. This advantage over heuristics using a fixed search radius, however, comes at the cost of using non-optimal step sizes also in those regimes in which the optimal rate is stable for a long time. This downside results in significant performance losses for many standard benchmark problems.

We introduce in this work a simple way to interpolate between the random global search of EAs and their deterministic counterparts which sample from a fixed radius only. To this end, we introduce \emph{normalized standard bit mutation}, in which the binomial choice of the search radius is replaced by a normal distribution. Normalized standard bit mutation allows a straightforward way to control its variance, and hence the degree of randomness involved. We experiment with a self-adjusting choice of this variance, and demonstrate its effectiveness for the two classic benchmark problems LeadingOnes and OneMax. Our work thereby also touches a largely ignored question in discrete evolutionary computation: multi-dimensional parameter control.
\end{abstract}


\section{Introduction}
\label{sec:intro}

Among the most successfully applied iterative optimization heuristics are local search variants and evolutionary algorithms (EAs). While the former sample at a fixed radius around previously evaluated solutions, most evolutionary algorithms classify as \emph{global search algorithms} which can escape local optima by creating offspring at larger distances. In the context of optimizing pseudo-Boolean functions $f:\{0,1\}^n \to \R$, for example, the most commonly found variation operator in EAs is \emph{standard bit mutation}. Standard bit mutation creates a new solution $y$ by flipping each bit of the parent individual $x \in \{0,1\}^n$ with some probability $0<p<1$, independently for each position. The probability to sample a specific offspring $y$ at distance $0 \le d \le n$ from $x$ thus equals $p^{H(x,y)}(1-p)^{H(x,y)}$, where $H(x,y)=| \{ 1 \le i \le n \mid x_i \neq y_i \}|$ denotes the Hamming distance of $x$ and $y$. This probability is strictly positive for all $y$, thus showing that the probability that an EA using standard bit mutation will have sampled a global optimum of $f$ converges to one as the number of iterations increases. In contrast to pure random search, however, the distance at which the offspring $y$ is sampled follows a binomial distribution, $\Bin(n,p)$, and is thus concentrated around its mean $np$. 

The ability to escape local optima comes at the price of frequent uses of non-optimal search radii even in those regimes in which the latter are stable for a long time. The incapability of standard bit mutation to adjust to such situations results in important performance losses on almost all classical benchmark functions, which often exhibit large parts of the optimization process in which flipping a certain number of bits is required. A convenient way to control the degree of randomness in the choice of the search radius would therefore be highly desirable. 

In this work we introduce such an interpolation. It allows to calibrate between deterministic and pure random search, while encompassing standard bit mutation as one specification. More precisely, we investigate \emph{normalized standard bit mutation}, in which the mutation strength (i.e., the search radius) is sampled from a normal distribution $N(\mu,\sigma^2)$. By choosing $\sigma=0$ one obtains a deterministic choice, and the ``degree of randomness'' increases with increasing $\sigma$. By the central limit theorem, we recover a distribution that is very similar to that of standard bit mutation by setting $\mu=np$ and $\sigma^2=np(1-p)$.

Apart from conceptual advantages, normalized standard bit mutation offers the advantage of separating the variance from the mean, which makes it easy to control both parameters independently during the optimization process. While multi-dimensional parameter control for discrete EAs is still in its infancy, cf. comments in~\cite{KarafotiasHE15,DoerrD18chapter}, we demonstrate in this work a simple, yet efficient way to control mean and variance of normalized standard bit mutation. 
As test case to investigate the benefits of normalized standard bit mutation we have chosen the 2-rate $(1+\lambda)$~EA$_{r/2,2r}$ from~\cite{DoerrGWY17}. The choice of this reference algorithm is based on our previous work~\cite{GECCO18} in which we observed, via a detailed fixed-target analysis of several $(1+\lambda)$~EAs, that for the two benchmark problems \onemax and \leadingones this algorithm performs significantly better than the plain $(1+\lambda)$~EA for a large range of initial target values. For both functions flipping one bit is optimal for a large fraction of the optimization process, cf. Figure~\ref{fig:optrates}. In these regimes the 2-rate $(1+\lambda)$~EA$_{r/2,2r}$ drastically looses performance due to sampling half the offspring with a mutation rate that is four times as large as the optimal one. Controlling the variance of this distribution seems therefore promising. 

On the way towards a $(1+\lambda)$~EA$_{r/2,2r}$ variant with self-adjusting choice of mean and variance we discover that already replacing the 2-rate sampling strategy of this algorithm by a normalized choice of the mutation strength significantly improves its performance. Controlling the variance then yields additional performance gains on the tested \onemax instances (we consider problem dimensions up to $10\,000$). On \leadingones, the variance control improves performance for small values of $\lambda$. Unlike one might first expect, for this test function the average optimization time (i.e., number of search points evaluated until an optimal solution is evaluated for the first time) of the $(1+50)$~variants of the $(1+\lambda)$~EA$_{r/2,2r}$ is better than that of their $(1+2)$ counterparts, which is an observation of independent interest.

\subsection{Related Work}
\label{sec:related}

We are not aware of any other work replacing the binomial search radius distribution of standard bit mutation by a normal distribution. We are also not aware of any work directly controlling the variance of the mutation strength distribution. As mentioned above, controlling more than one parameter simultaneously is a largely ignored question in discrete evolutionary computation (EC).

A recently developed algorithm that also addresses the idea to sample the search radius from a different distribution than the binomial one is the fast-GA introduced in~\cite{DoerrLMN17}. It samples the mutation strength from a power-law distribution, thus essentially shifting probability mass from small mutation strengths to larger ones. It is shown in~\cite{DoerrLMN17} that the fast-GA is very efficient on so-called \textsc{Jump}$_m$ functions, which require to flip $m$ bits simultaneously to jump from a local to the global optimum. It is furthermore discussed in~\cite{DoerrLMN17} that the advantages of the fast-GA do not sacrifice too drastically the performance on uni-modal benchmark functions such as \onemax and \leadingones. This work has already received considerable attention in the literature~\cite{MironovichB17,FriedrichQW18,FriedrichGQW18,CorusOY18a,Lengler18}. However, only static distributions are considered so far, and it is very likely that a control mechanism similar to the ones proposed in this work would be beneficial. We will comment on this in Section~\ref{sec:conclusions}.  


As reasoned above, normalized standard bit mutation offers an elegant way to interpolate between deterministic mutation strengths and regular standard bit mutation, thus showing that Randomized Local Search (RLS) variants with their deterministic search radii and the (1+1) EA with mutation rate $p$ are essentially just different instantiations of the same meta-algorithm. Similar results also extend to population-based $(\mu+\lambda)$~EAs. Note that normalized standard bit mutation also allows other degrees of randomization, thereby offering a wide range for further experimentation. In this context we note that for the special case of standard RLS (i.e., the greedy (1+1) hill climber that flips in each iteration exactly one uniformly chosen bit) a similar meta-model allowing to interpolate between the (1+1)~EA and RLS is the (1+1)~EA$_{>0}$ introduced in~\cite{JansenZ11,CarvalhoD18}. This model, however, is much less flexible, and does not allow, for example, deterministic search radii greater than one.

\subsection{Experimental Setup}
\label{sec:setup}

Unless stated otherwise, all numbers reported in this work are based on 100 independent runs of the respective algorithms. To ease readability, we only display average values. All raw data as well as detailed summaries with quantiles, standard deviations, etc. are available at \url{https://github.com/FurongYe/Fixed-Target-Results}. Selected statistical results can be found in Tables~\ref{fig:OMstats} and~\ref{fig:LOstats}, respectively. These summaries have been created with IOHprofiler, our recently announced benchmarking and data analysis tool~\cite{IOHprofiler}. 

\section{Previous Observations for the Two-Rate \texorpdfstring{$(1+\lambda)$~EA}{EA} and the Two Benchmark Problems}
\label{sec:prelims}

A starting point of our work are results presented in~\cite{GECCO18}. In this work we observed that the evolutionary algorithm with success-based self-adjusting mutation rate proposed in~\cite{DoerrGWY17} outperforms the \opl for a large range of sub-optimal targets. It then drastically looses performance in the later parts of the optimization process, which results in an overall poor optimization time on \onemax and \leadingones functions of moderate problem dimensions $n \le 10\,000$. The in~\cite{DoerrGWY17} proven optimal asymptotic behavior on \onemax can thus not be observed for these dimensions. 

We briefly summarize in this section the algorithm from~\cite{DoerrGWY17} and the results presented in~\cite{GECCO18}. We also discuss a few basic properties of the two benchmark problems, which explain the choices made in subsequent sections.

\subsection{The Two-Rate EA}
\label{sec:basic}

\begin{figure*}[t]
\centering
\includegraphics[width=\linewidth]{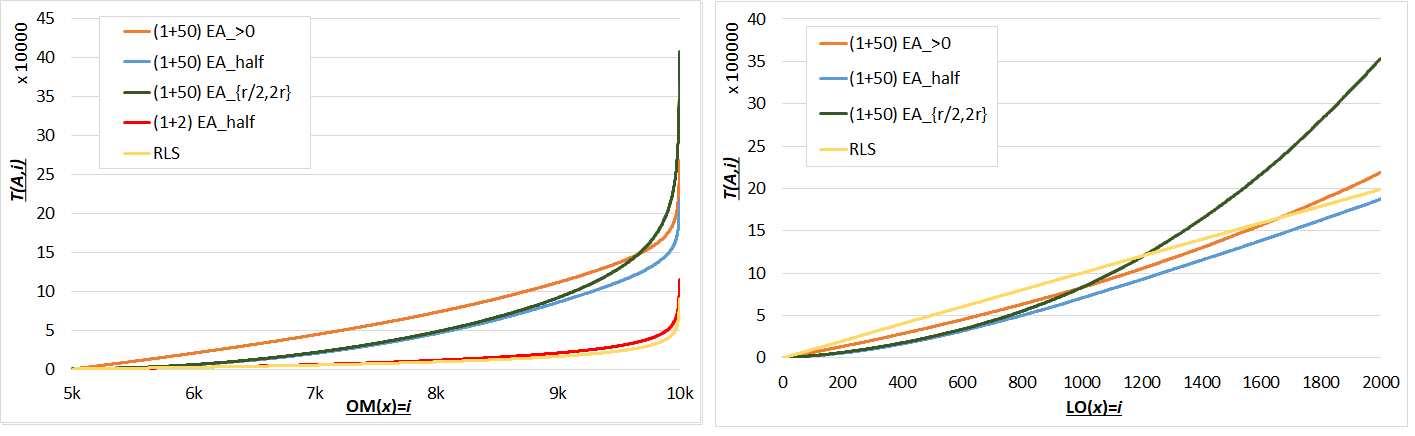}
\caption{Average fixed-target running times for variants of the 2-rate $(1+50)$~EA for $10\,000$-dimensional \onemax and $2\,000$-dimensional \leadingones.}
\label{fig:2rate}
\end{figure*}

The algorithm introduced in~\cite{DoerrGWY17}, which we named $(1+\lambda)$~EA$_{r/2,2r}$ in~\cite{GECCO18}, is a \opl which applies in each iteration two different mutation rates. Half of the offspring population is generated with mutation rate $r/(2n)$, the other half with mutation rate $2r/n$. The parameter $r$ is the current best mutation strength, which is updated after each iteration, with a bias towards the rate by which the best of the $\lambda$ offspring has been sampled, cf. Algorithm~\ref{alg:DoerrGWY17} for details. 

 \begin{algorithm2e}%
	\textbf{Initialization:} 
	Sample $x \in \{0,1\}^{n}$ uniformly at random and evaluate $f(x)$\;
	Initialize $r \assign r^{\text{init}}$; // Following~\cite{DoerrGWY17} we use $r^{\text{init}}=2$\;
  \textbf{Optimization:}
	\For{$t=1,2,3,\ldots$}{
		\For{$i=1,\ldots,\lambda/2$}{
			\label{line:half}Sample $\ell^{(i)} \sim \Bin_{>0}(n,r/(2n))$, 
			create $y^{(i)} \assign \mut_{\ell^{(i)}}(x)$, and
			evaluate $f(y^{(i)})$\;
		 }
		\For{$i=\lambda/2+1,\ldots,\lambda$}{
			\label{line:double}Sample $\ell^{(i)} \sim \Bin_{>0}(n,2r/n)$, 
			create $y^{(i)} \assign \mut_{\ell^{(i)}}(x)$, and
			evaluate $f(y^{(i)})$\;
		 }
		$x^* \assign \arg\max\{f(y^{(1)}), \ldots, f(y^{(\lambda)})\}$ (ties broken u.a.r.)\;
		\lIf{$f(x^*)\ge f(x)$}{$x \assign x^*$}
		\lIf{$x^{*}$ has been created with mutation rate $r/2$}{$s \assign 3/4$ \textbf{ else} $s \assign 1/4$}
		Sample $q \in [0,1]$ u.a.r.\;
		\lIf{$q\le s$}{$r \assign \max\{r/2,2\}$ \textbf{ else} {$r \assign \min\{2r,n/4\}$}} 
	}
\caption{The 2-rate $(1+\lambda)$~EA$_{r/2,2r}$ with adaptive mutation rates proposed in~\cite{DoerrGWY17}}
\label{alg:DoerrGWY17}
\end{algorithm2e}  

Note that here and in the following we make use of the fact that standard bit mutation, which is traditionally defined by flipping each bit in a length-$n$ bit string with some probability $p$ (independently of all other decisions), can be equivalently described by first sampling a radius $\ell$ from the binomial distribution $\Bin(n,p)$ and then applying the $\mut_{\ell}$ operator, which flips $\ell$ pairwise different bits that are chosen from the index set $[n]=\{1,2,\ldots,n\}$ uniformly at random. 

Following the discussions and the notation introduced in~\cite{CarvalhoD18,DoerrW18,GECCO18} we enforce in this work that all offspring differ from their parents by at least one bit. We therefore require in lines~\ref{line:half} and~\ref{line:double} that the \emph{mutation strength} $\ell$ is at least one. This is achieved by re-sampling if needed, or, equivalently, by sampling from the conditional binomial distribution $\Bin_{>0}(n,p)$ which assigns to each value $k \in [n]$ a probability of $\Bin(n,p)(k)/(1-(1-p)^n)=\binom{n}{k}p^k(1-p)^{n-k}/(1-(1-p)^n)$. 

In~\cite{GECCO18} we compared the fixed-target performance of the $(1+50)$~EA$_{>0}$ (i.e., the \opl using the conditional sampling rule introduced above) and the $(1+50)$~EA$_{r/2,2r}$ on \onemax and \leadingones. These two classic optimization problems ask to maximize the functions $\{0,1\}^n \rightarrow \{0\} \cup [n]$ which are defined via
$\textsc{OneMax}(x) = \sum_{i=1}^n{x_i} \text{ and}$
$\textsc{LeadingOnes}(x) = \max \{i \in [0..n] \mid \forall j \leq i: x_j = 1\},$  
respectively. In Figure~\ref{fig:2rate} we report similar empirical results for $n=10\,000$ (\onemax) and $n=2\,000$ (\leadingones) (the other results in the two figures will be addressed below). We observed in~\cite{GECCO18} that for both functions the $(1+50)$~EA$_{r/2,2r}$ from~\cite{DoerrGWY17} performs well for small target values, but drastically looses performance in the later stages of the optimization process. 

\subsection{Properties of the Benchmark Problems}
\label{sec:analyzingFTP}

\begin{figure*}[t]
\centering
\includegraphics[width=\linewidth]{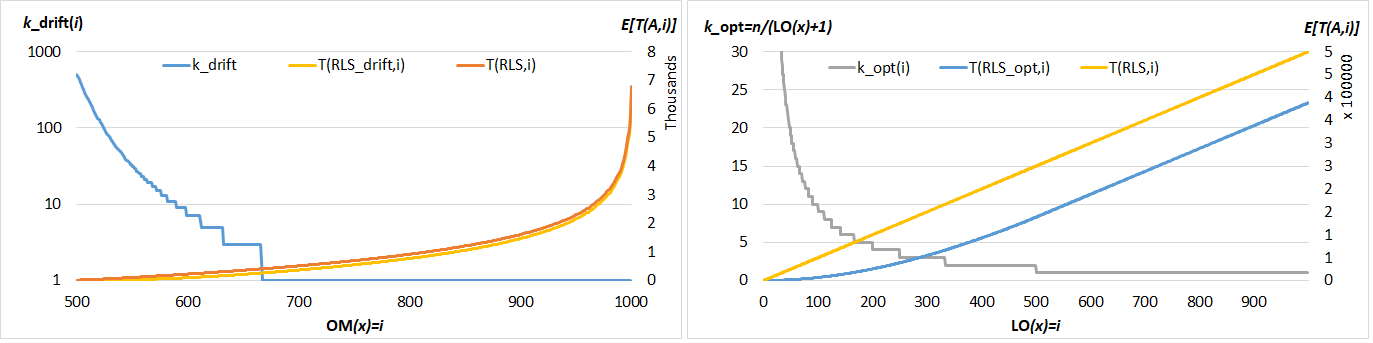}
\caption{Drift maximizing and optimal mutation strength for 1000-dimensional \onemax and \leadingones functions, respectively. Note the logarithmic scale for $k_\text{drift}$ for \onemax. For \onemax, RLS spends around $94\%$ of the total optimization time in the regime in which $k_{\text{drift}}=1$, for $\leadingones$ this fraction is still $50\%$. For the drift-maximizing/optimal RLS-variants  flipping in each iteration $k_{\text{drift}}$ and $k_{\text{opt}}$ bits, respectively, these fractions are around $96\%$ for \onemax and $64\%$ for \leadingones.}
\label{fig:optrates}
\end{figure*}

Both \onemax and \leadingones have a long period during the optimization run in which flipping one bit is optimal. 

For \onemax flipping one bit is widely assumed to be optimal as soon as $f(x)\ge 2n/3$. Quite interestingly, however, this conjecture has not not been rigorously proven to date. It is only known that drift-maximizing mutation strengths are \emph{almost} optimal~\cite{DoerrDY16}, in the sense that the overall expected optimization time of the elitist (1+1) algorithm using these rates in each step cannot be worse than the best-possible unary unbiased algorithm for \onemax by more than an additive $o(n)$ lower order term~\cite{DoerrDY16}. But even for the drift maximizer the statement that flipping one bit is optimal when $f(x)\ge 2n/3$ has only be shown for an approximation, not the actual drift maximizer. Numerical evaluations for problem dimensions up to $10\,000$ nevertheless confirm that 1-bit flips are optimal when the \onemax-value exceeds $2n/3$. 

For \leadingones, on the other hand, it is well known that flipping one bit is optimal as soon as $f(x)\ge n/2$~\cite{Doerr18LO}.

We display in Figure~\ref{fig:optrates}, which is adjusted from~\cite{Doerr18tutorial}, the optimal and drift-maximizing mutation strength for \leadingones and \onemax, respectively. We also display in the same figure the expected time needed by RLS$_{\text{opt}}$ and RLS$_{\text{drift}}$, the elitist (1+1) algorithm using in each step these mutation rates. We see that these algorithms spend around $96\%$ (for \onemax) and $64\%$ (for \leadingones), respectively, of their time in the regime where flipping one bit is (almost) optimal. These numbers are based on an exact computation for $\leadingones$ and on an empirical evaluation of $500$ independent runs for \onemax.

\subsection{Implications for the \texorpdfstring{$(1+50)$~EA$_{r/2,2r}$}{(1+50) EA}}
\label{sec:implications}

Assume that in the regime of optimal one-bit flips the $(1+50)$~EA$_{r/2,2r}$ has correctly identified that flipping one bit is optimal. It will hence use the smallest possible value for $r$, which is $2$. In this case, half the offspring are sampled with (the for this algorithm optimal) mutation rate $1/n$, while the other half of the offspring population is sampled with mutation rate $4/n$, thus flipping on average more than four times the optimal number of bits. It is therefore non-surprising that in this regime (and already before) the gradient of the average fixed-target running time curves in Figures~\ref{fig:2rate} are much worse for the $(1+50)$~EA$_{r/2,2r}$ than for the $(1+50)$~EA$_{>0}$.

\section{Creating Half the Offspring with Optimal Mutation Rate}
\label{sec:EA_MS}  

The observations made in the last section inspire our first algorithms, the $(1+\lambda)$~EA$_{r,U(0,\sigma r/n)}$ defined via Algorithm~\ref{alg:sigma}. This algorithm samples half the offspring using as deterministic mutation strength the best mutation strength of the last iteration. The other offspring are sampled with a mutation rate that is sampled uniformly at random from the interval $(0,\sigma r/n)$. 

\begin{algorithm2e}%
	\textbf{Initialization:} 
	Sample $x \in \{0,1\}^{n}$ uniformly at random and evaluate $f(x)$\;
	Initialize $r \assign r^{\text{init}}$; // we use $r^{\text{init}}=2$\;
  \textbf{Optimization:}
	\For{$t=1,2,3,\ldots$}{
		\For{$i=1,\ldots,\lambda/2$}{
			\label{line:halfsigma}Set $\ell^{(i)} \assign r$, 
			create $y^{(i)} \assign \mut_{\ell^{(i)}}(x)$, and
			evaluate $f(y^{(i)})$\;
		 }
		\For{$i=\lambda/2+1,\ldots,\lambda$}{
			\label{line:doublesigma}Sample $p^{(i)} \sim \min\{U(0,\sigma r/n),1\}$, 
			$\ell^{(i)} \sim \Bin_{>0}(n,p^{(i)})$, 
			create $y^{(i)} \assign \mut_{\ell^{(i)}}(x)$, and
			evaluate $f(y^{(i)})$\;
		 }
		$i \assign \min\left\{ j \mid f(y^{(j)}) = \max\{f(y^{(k)}) \mid k \in [n]\} \right\}$\;
				$r \assign \ell^{(i)}$\; 
		\lIf{$f(y^{(i)})\ge f(x)$}{$x \assign y^{(i)}$}
	}
\caption{The $(1+\lambda)$~EA$_{r,U(0,\sigma r/n)}$. In line~\ref{line:doublesigma} we denote by $U(a,b)$ the uniform distribution in the interval $(a,b)$. For $\sigma=2$ we call this algorithm the $(1+\lambda)$~EA$_{\text{half}}$.}
\label{alg:sigma}
\end{algorithm2e} 

\begin{figure}[t]
\centering
\includegraphics[width=\linewidth]{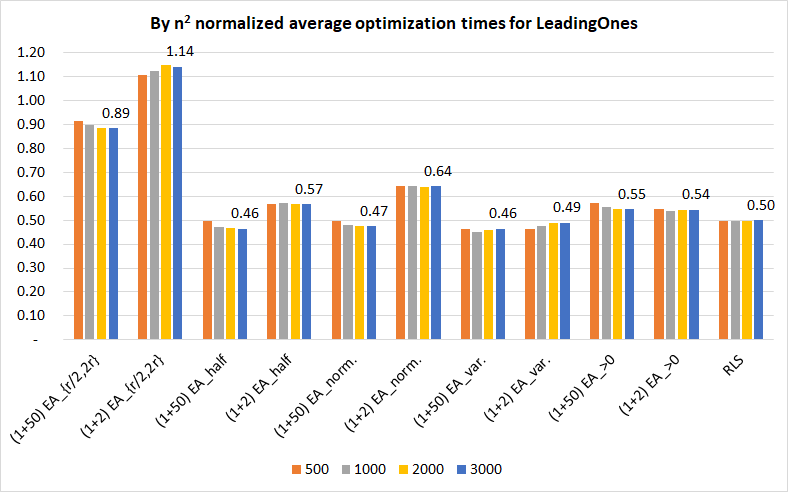}
\caption{By $n^2$ normalized average optimization times for \leadingones, for $n$ between 500 and $3\,000$. Displayed numbers are for $n=3\,000$.}
\label{fig:summaryLO}
\end{figure}

As we can see in Figure~\ref{fig:2rate} this algorithm significantly improves the performance in those later parts of the optimization process. Normalized total optimization times for various problem dimensions are provided in Figures~\ref{fig:summaryOM} and~\ref{fig:summaryLO}, respectively. We display data for $\sigma=2$ only, and call this $(1+\lambda)$~EA$_{r,U(0,\sigma r/n)}$ variant $(1+\lambda)$~EA$_{\text{half}}$. We note that smaller values of $\sigma$, e.g., $\sigma=1.5$ would give better results. The same effect would be observable when replacing the factor two in the $(1+\lambda)$~EA$_{r/(2n),2r}$, i.e., when using a $(1+\lambda)$~EA$_{r/(\sigma n),\sigma r}$ rule with $\sigma \neq 2$. A detailed discussion of this effect is omitted here for reasons of space. 

It is remarkable that on \leadingones the $(1+\lambda)$~EA$_{\text{half}}$ performs better than Randomized Local Search (RLS), the elitist (1+1) algorithm flipping in each iteration exactly one uniformly chosen bit. The slightly worse gradients for target values $v>n/2$ (which are a consequence of randomly sampling the mutation rate instead of using mutation strength one deterministically) are compensated for by the gains made in the initial phase of the optimization process, where the EA variants benefit from larger mutation rates. 

On \onemax the performance of the $(1+\lambda)$~EA$_{\text{half}}$ is better than that of the plain $(1+\lambda)$~EA$_{>0}$ for both tested values $\lambda=50$ and $\lambda=2$.

We recall that it is well known that, both for \onemax and \leadingones, the optimal offspring population size in the regular $(1+\lambda)$~EA is $\lambda=1$~\cite{JansenJW05}. A monotonic dependence of the average optimization time on $\lambda$ is conjectured (and empirically observed) but not formally proven. While for \onemax the impact of $\lambda$ is significant, the dependency on $\lambda$ is much less pronounced for \leadingones. Empirical results for both functions and a theoretical running time analysis for \leadingones can be found in~\cite{GECCO18}. For \onemax \cite{GiessenW17Algorithmica} offers a precise running time analysis of the \opl for broad ranges of offspring population sizes $\lambda$ and mutation rates $p=c/n$. 
In light of the fact that the theoretical considerations in~\cite{DoerrGWY17} required $\lambda = \omega(1)$, it is worthwhile to note that for all tested problem dimensions the $(1+2)$~EA$_{r/2,2r}$ performs better on \onemax than the $(1+50)$~EA$_{r/2,2r}$. Note, however, that the inverse holds for \leadingones, cf. Figure~\ref{fig:summaryLO}. For this function it seems to be important that the number of offspring allows a better estimation of the better mutation rate. We will observe the same phenomenon for all other algorithms introduced below.

\section{Normalized Standard Bit Mutation}
\label{sec:normalized}

\begin{figure}[t]
\centering
\includegraphics[width=\linewidth]{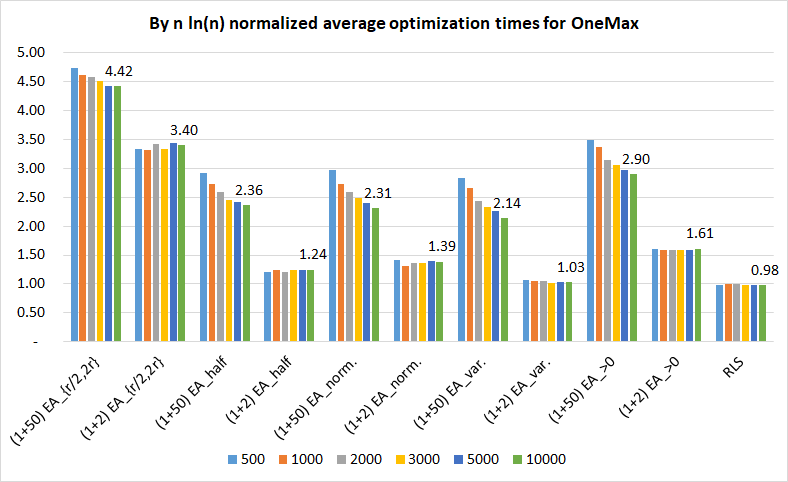}
\caption{By $n \ln(n)$ normalized average optimization times for \onemax, for $n$ between 500 and $10\,000$. Displayed numbers are for $n=10\,000$.}
\label{fig:summaryOM}
\end{figure}

In light of the results presented in the previous section, one may wonder if splitting the population into two halves is needed after all. We investigate this question by introducing the $(1+\lambda)$~EA$_{\text{norm.}}$ which in each iteration and for each $i \in [\lambda]$ samples the mutation strength $\ell^{(i)}$ from the normal distribution $N(r,r(1-r/n))$ around the best mutation strength $r$ of the previous iteration and rounding the sampled value to the closest integer. The reasons to replace the uniform distribution $U(r/n-\sigma,r/n+\sigma)$ will be addressed below. As before we enforce $\ell^{(i)} \ge 1$ by re-sampling if needed, thus effectively sampling the mutation strength from the conditional distribution $N_{>0}(r,r(1-r/n))$. Algorithm~\ref{alg:normal} summarizes this algorithm. 

 \begin{algorithm2e}%
	\textbf{Initialization:} 
	Sample $x \in \{0,1\}^{n}$ uniformly at random and evaluate $f(x)$\;
	Initialize $r \assign r^{\text{init}}$; // we use $r^{\text{init}}=2$\;
  \textbf{Optimization:}
	\For{$t=1,2,3,\ldots$}{
		\For{$i=1,\ldots,\lambda$}{
			\label{line:mutnormal}Sample 
			$\ell^{(i)} \sim \min\{N_{>0}(r,r(1-r/n)),n\}$, 
			create $y^{(i)} \assign \mut_{\ell^{(i)}}(x)$, and
			evaluate $f(y^{(i)})$\;
		 }
		$i \assign \min\left\{ j \mid f(y^{(j)}) = \max\{f(y^{(k)}) \mid k \in [n]\} \right\}$\;
				$r \assign \ell^{(i)}$\; 
		\lIf{$f(y^{(i)})\ge f(x)$}{$x \assign y^{(i)}$}
	}
\caption{The $(1+\lambda)$~EA$_{\text{norm.}}$ with normalized standard bit mutation}
\label{alg:normal}
\end{algorithm2e} 

Note that the variance $r(1-r/n)$ of the unconditional normal distribution $N(r,r(1-r/n))$ is identical to that of the unconditional binomial distribution $\Bin(n,r/n)$. We use the normal distribution here for reasons that will be explained in the next section. Note, however, that very similar results would be obtained when replacing in line~\ref{line:mutnormal} of Algorithm~\ref{alg:normal} the normal distribution $N_{>0}(r,r(1-r/n))$ by the binomial one $\Bin_{>0}(n,r/n)$. We briefly recall that, by the central limit theorem, the (unconditional) binomial distribution converges to the (unconditional) normal distribution. 

The empirical performance of the $(1+50)$~EA$_{\text{norm.}}$ is comparable to that of the $(1+50)$~EA$_{\text{half}}$ for both problems and all tested problem dimensions, cf. Figures~\ref{fig:summaryOM} and~\ref{fig:summaryLO}. Note, however, that for $\lambda=2$ the $(1+2)$~EA$_{\text{norm.}}$ performs worse than the $(1+2)$~EA$_{\text{half}}$.

\subsection{Interpolating Local and Global Search}
\label{sec:variance}

As discussed above, all EA variants mentioned so far suffer from the variance of the random selection of the mutation rate, in particular in the long final part of the optimization process in which the optimal mutation strength is one. We therefore analyze a simple way to reduce this variance on the fly. To this end, we build upon the $(1+\lambda)$~EA$_{\text{norm.}}$ and introduce a counter $c$, which is initialized at zero. In each iteration, we check if the value of $r$ changes. If so, the counter is re-set to zero. It is increased by one otherwise, i.e., if the value of $r$ remains the same. We use this counter to self-adjust the variance of the normal distribution. To this end, we replace in line~\ref{line:mutnormal} of Algorithm~\ref{alg:normal} the conditional normal distribution $N_{>0}(r,r(1-r/n))$ by the conditional normal distribution $N_{>0}(r,F^c r(1-r/n))$, where $F<1$ is a constant discount factor. Algorithm~\ref{alg:normalada} summarizes this \opl variant with normalized standard bit mutation and a self-adjusting choice of mean and variance. 

 \begin{algorithm2e}%
	\textbf{Initialization:} 
	Sample $x \in \{0,1\}^{n}$ uniformly at random and evaluate $f(x)$\;
	Initialize $r \assign r^{\text{init}}$; // we use $r^{\text{init}}=2$\;
	Initialize $c \assign 0$\;
  \textbf{Optimization:}
	\For{$t=1,2,3,\ldots$}{
		\For{$i=1,\ldots,\lambda$}{
			\label{line:mutnormaladap}Sample 
			$\ell^{(i)} \sim \min\{N_{>0}(r,F^c r(1-r/n)),n\}$, 
			create $y^{(i)} \assign \mut_{\ell^{(i)}}(x)$, and
			evaluate $f(y^{(i)})$\;
		 }
		$i \assign \min\left\{ j \mid f(y^{(j)}) = \max\{f(y^{(k)}) \mid k \in [n]\} \right\}$\;
		\lIf{$r =\ell^{(i)}$}{$c \assign c+1$; \textbf{else} $c \assign 0$}
		$r \assign \ell^{(i)}$\; 
		\lIf{$f(y^{(i)})\ge f(x)$}{$x \assign y^{(i)}$}
	}
\caption{The $(1+\lambda)$~EA$_{\text{var.}}$ with normalized standard bit mutation and a self-adjusting choice of mean and variance}
\label{alg:normalada}
\end{algorithm2e}

\textbf{Choice of $F$:} We use $F=0.98$ in all reported experiments. Preliminary tests suggest that values $F<0.95$ are not advisable, since the algorithm may get stuck with sub-optimal mutation rates. This could be avoided by introducing a lower bound for the variance and/or by mechanisms taking into account whether or not an iteration has been \emph{successful}, i.e., whether it has produced a strictly better offspring. 

\begin{figure*}[t]
\centering
\includegraphics[width=\linewidth]{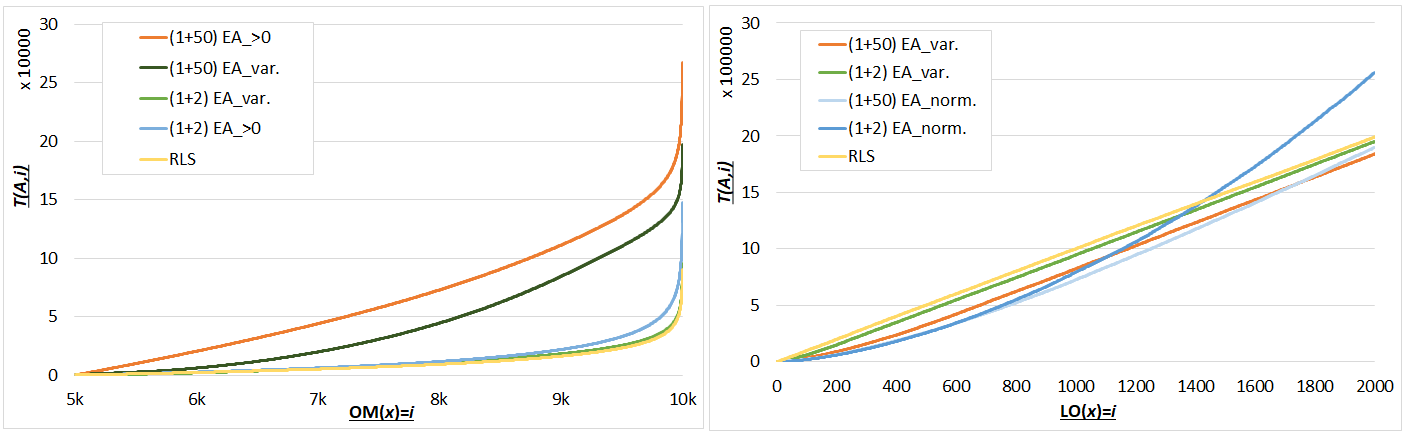}
\caption{Average fixed-target running times for $10\,000$-dimensional \onemax and $2\,000$-dimensional \leadingones.}
\label{fig:2rate2}
\end{figure*}

The empirical comparison suggests that the self-adjusting choice of the variance in the $(1+\lambda)$~EA$_{\text{var.}}$ improves the performance on \onemax further, cf. also Figure~\ref{fig:2rate2} for average fixed-target results for $n=10\,000$. For $\lambda=2$ the average performance is comparable to, but slightly worse than that of RLS. For \leadingones, the $(1+50)$~EA$_{\text{var.}}$ is comparable in performance to the $(1+50)$~EA$_{\text{norm.}}$, but we observe that for $\lambda=2$ the $(1+\lambda)$~EA$_{\text{var.}}$ performs better. It is the only one among all tested EAs for which decreasing $\lambda$ from 50 to 2 does not result in a significantly increased running time.

\section{A Meta-Algorithm with Normalized Standard Bit Mutation}
\label{sec:meta}

In the $(1+\lambda)$~EA$_{\text{var.}}$ we make use of the fact that a small variance in line~\ref{line:mutnormaladap} of Algorithm~\ref{alg:normalada} results in a more concentrated distribution. The variance adjustment is thus an efficient way to steer the \emph{degree of randomness} in the selection of the mutation rate. It allows to interpolate between deterministic and random mutation rates. In our experimentation we do not go beyond the variance of the binomial distribution, but in principle there is no reason to not regard larger variance as well. The question of how to best determine the degree of randomness in the choice of the mutation rate has, to the best of our knowledge, not previously been addressed in the EC literature. We believe that this idea carries good potential, since it demonstrates that local search with its deterministic search radius and evolutionary algorithms with their global search radii are merely two different configurations of the same meta-algorithm, and not two different algorithms as the general perception might indicate. To make this point very explicit, we introduce with Algorithm~\ref{alg:meta} a general meta-algorithm, of which local search with deterministic mutation strengths and EAs are special instantiations. 

Note that in this meta-model we use static parameter values, variants with adaptive mutation rates can be obtained by applying the usual parameter control techniques, as demonstrated above. Of course, the same normalization can be done for similar EAs, the technique is not restricted to elitist $(1+\lambda)$-type algorithms. Likewise, the condition to flip at least one bit can be omitted, i.e., one can replace the conditional normal distribution $N_{>0}(r,\sigma^2)$ in line~\ref{line:mutmeta} by the unconditional $N(r,\sigma^2)$. 

 \begin{algorithm2e}%
	\textbf{Initialization:} 
	Sample $x \in \{0,1\}^{n}$ uniformly at random and evaluate $f(x)$\;
  \textbf{Optimization:}
	\For{$t=1,2,3,\ldots$}{
		\For{$i=1,\ldots,\lambda$}{
			\label{line:mutmeta}
			Sample $\ell^{(i)} \sim \min\{N_{>0}(r,\sigma^2),n\}$, 
			create $y^{(i)} \assign \mut_{\ell^{(i)}}(x)$, and
			evaluate $f(y^{(i)})$\;
		 }
		$y \assign \arg\max\{f(y^{(k)}) \mid k \in [n]\}$\;
		\lIf{$f(y)\ge f(x)$}{$x \assign y$}
	}
\caption{The $(1+\lambda)$~Meta-Algorithm with (static) normalized standard bit mutation. The RLS variant with deterministic search radius $r$ and \opl using standard bit mutation with mutation rate $r/n$ are identical to this algorithm with $\sigma^2=0$ and $\sigma^2=r(1-r/n)$, respectively.}
\label{alg:meta}
\end{algorithm2e} 

\section{Discussion and Outlook}
\label{sec:conclusions}

We have introduced in this work \emph{normalized standard bit mutation,} which replaces the binomial choice of the mutation strength in standard bit mutation by a normal distribution. This normalization allows a straightforward way to control the variance of the distribution, which can now be adjusted independently of the mean. We have demonstrated that such an approach can be beneficial when optimizing classic benchmark problems such as \leadingones and \onemax. In future work, we plan to validate our approach for the fast-GA proposed in~\cite{DoerrLMN17}. We are confident that variance control should be beneficial for that algorithm as well.  

Our work has concentrated on \onemax and \leadingones, as two examples where the optimal mutation rate is stable for a long time. When applied in practice---where abrupt changes of the optimal mutation strengths may occur---our variance control mechanism needs to be modified so that the variance is increased if no strict progress has been observed for a sufficiently long period. We plan to investigate this question by studying concatenated jump-functions, i.e., functions for which one mutation strength is optimal for some significant number of iterations, followed by a situation in which a much larger number of bits need to be flipped in order to make progress. 

Related to the point made in the last paragraph, we also note that the parameter control technique which we applied to adjust the mean of the sampling distribution for the mutation strength has an extremely short learning period, since we simply use the best mutation strength of the last iteration as mean for the sampling distribution of the next iteration. For more rugged fitness landscapes a proper learning, which takes into account several iterations, should be preferable.

We recall that multi-dimensional parameter control has not received much attention in the EC literature for discrete optimization problems~\cite{KarafotiasHE15,DoerrD18chapter}. Our work falls into this category, and we have demonstrated a simple way to separate the control of the mean from that of the variance of the mutation strength distribution. We hope that our work inspires more research in this direction, since practical EAs tend to have many different parameters that need to be adjusted during the optimization process. 

Finally, another avenue for further work is provided by the meta-algorithm presented in Section~\ref{sec:meta}, which demonstrates that Randomized Local Search and evolutionary algorithms can be seen as two configurations of the meta-algorithm. Parameter control, or, in this context possibly more suitably referred to as \emph{online algorithm configuration}, offers the possibility to interpolate between these algorithms (and even more drastically randomized heuristics). Given the significant advances in the context of \emph{algorithm configuration} witnessed by the EC and machine learning communities, we believe that such meta-models carry significant potential to exploit and profit from advantages of different heuristics. Note here that the configuration of meta-algorithms offers much more flexibility than the \emph{algorithm selection} approach classically taken in EC, e.g., in most works on hyper-heuristics.

\newcommand{\etalchar}[1]{$^{#1}$}
\providecommand{\bysame}{\leavevmode\hbox to3em{\hrulefill}\thinspace}
\providecommand{\MR}{\relax\ifhmode\unskip\space\fi MR }
\providecommand{\MRhref}[2]{%
  \href{http://www.ams.org/mathscinet-getitem?mr=#1}{#2}
}
\providecommand{\href}[2]{#2}


\begin{landscape}
\begin{table}
\begin{tabular}{lrrrrrrrrrr}
&&&& \multicolumn{7}{c}{quantiles}\\
\cmidrule(lr){5-11}
 Algorithm                  & target value & AHT & rsd & 2\%     & 10\%    & 25\%    & 50\%    & 75\%    & 90\%    & 98\%    \\
\hline
(1+50) EA\_\{r/2,2r\}      & 6,666   & 15,044  & 4\%         & 13,902  & 14,371  & 14,588  & 15,000  & 15,447  & 15,812  & 16,254  \\
(1+2) EA\_\{r/2,2r\}       & 6,666   & 3,870   & 4\%         & 3,548   & 3,648   & 3,740   & 3,883   & 3,968   & 4,063   & 4,184   \\
(1+50) EA\_half            & 6,666   & 14,066  & 4\%         & 12,955  & 13,403  & 13,578  & 14,004  & 14,462  & 14,901  & 15,302  \\
(1+2) EA\_half             & 6,666   & 4,398   & 4\%         & 4,077   & 4,204   & 4,294   & 4,397   & 4,518   & 4,586   & 4,682   \\
(1+50) EA\_norm.           & 6,666   & 14,121  & 5\%         & 12,753  & 13,216  & 13,541  & 14,055  & 14,631  & 15,141  & 15,802  \\
(1+2) EA\_norm.            & 6,666   & 4,113   & 5\%         & 3,633   & 3,814   & 3,970   & 4,115   & 4,268   & 4,365   & 4,549   \\
(1+50) EA\_var.            & 6,666   & 14,157  & 6\%         & 12,603  & 13,202  & 13,603  & 14,009  & 14,518  & 14,954  & 16,406  \\
(1+2) EA\_var.             & 6,666   & 4,101   & 6\%         & 3,539   & 3,794   & 3,908   & 4,090   & 4,292   & 4,409   & 4,593   \\
(1+50) EA\_\textgreater{}0 & 6,666   & 35,667  & 3\%         & 33,952  & 34,608  & 34,968  & 35,655  & 36,261  & 36,903  & 37,206  \\
(1+2) EA\_\textgreater{}0  & 6,666   & 4,803   & 3\%         & 4,528   & 4,615   & 4,704   & 4,785   & 4,910   & 4,994   & 5,059   \\
RLS                        & 6,666   & 4,074   & 3\%         & 3,866   & 3,922   & 3,981   & 4,056   & 4,181   & 4,243   & 4,282   \\
\hline 
(1+50) EA\_\{r/2,2r\}      & 10,000  & 407,229 & 11\%        & 338,465 & 358,460 & 374,563 & 400,705 & 424,652 & 460,563 & 532,973 \\
(1+2) EA\_\{r/2,2r\}       & 10,000  & 313,508 & 14\%        & 245,124 & 260,966 & 280,514 & 304,278 & 339,070 & 377,008 & 406,688 \\
(1+50) EA\_half            & 10,000  & 217,571 & 8\%         & 192,545 & 200,906 & 205,877 & 213,824 & 223,807 & 243,920 & 262,806 \\
(1+2) EA\_half             & 10,000  & 114,397 & 14\%        & 89,611  & 93,332  & 103,964 & 110,820 & 124,638 & 138,760 & 149,190 \\
(1+50) EA\_norm.           & 10,000  & 212,727 & 7\%         & 189,980 & 196,034 & 201,215 & 210,546 & 220,070 & 231,249 & 258,132 \\
(1+2) EA\_norm.            & 10,000  & 127,699 & 14\%        & 97,719  & 108,549 & 115,732 & 124,222 & 135,542 & 153,396 & 170,056 \\
(1+50) EA\_var.            & 10,000  & 196,912 & 7\%         & 176,463 & 181,106 & 186,208 & 195,329 & 205,624 & 213,984 & 223,440 \\
(1+2) EA\_var.             & 10,000  & 95,149  & 15\%        & 73,683  & 79,346  & 83,983  & 92,610  & 101,793 & 113,507 & 131,574 \\
(1+50) EA\_\textgreater{}0 & 10,000  & 267,400 & 7\%         & 239,422 & 245,132 & 253,204 & 263,883 & 278,150 & 293,031 & 304,119 \\
(1+2) EA\_\textgreater{}0  & 10,000  & 147,909 & 14\%        & 105,464 & 122,542 & 133,903 & 143,571 & 160,084 & 175,385 & 196,034 \\
RLS                        & 10,000  & 90,276  & 13\%        & 71,490  & 76,497  & 81,235  & 87,841  & 98,103  & 106,375 & 115,385\\
\hline \vspace{1ex}
\end{tabular}
\caption{Selected statistics of the fixed-target running times for the $n=10\,000$-dimensional \onemax problem. AHT= average first hitting time, rsd= relative standard deviation. We have chosen to display target value $2n/3=6\,666$ because this is the point after which flipping one bit becomes optimal, i.e., advantages over RLS must result from the phase before reaching this target point.}
\label{fig:OMstats}
\end{table}
\end{landscape}

\begin{landscape}
\begin{table}
\begin{tabular}{lrrrrrrrrrr}
&&&& \multicolumn{7}{c}{quantiles}\\
\cmidrule(lr){5-11}
 Algorithm                  & target value & AHT & rsd & 2\%     & 10\%    & 25\%    & 50\%    & 75\%    & 90\%    & 98\%    \\
\hline
(1+50) EA\_\{r/2,2r\}      & 1,000   & 831,517   & 6\%         & 713,071   & 758,260   & 797,275   & 837,006   & 865,670   & 894,654   & 912,238   \\
(1+2) EA\_\{r/2,2r\}       & 1,000   & 981,137   & 7\%         & 831,765   & 898,444   & 945,445   & 978,358   & 1,016,216 & 1,074,253 & 1,136,684 \\
(1+50) EA\_half            & 1,000   & 706,871   & 6\%         & 628,259   & 651,179   & 672,543   & 701,703   & 736,176   & 768,353   & 791,133   \\
(1+2) EA\_half             & 1,000   & 823,231   & 6\%         & 712,835   & 765,862   & 789,186   & 822,706   & 854,330   & 887,096   & 931,361   \\
(1+50) EA\_norm.           & 1,000   & 726,797   & 5\%         & 656,740   & 670,051   & 696,888   & 731,413   & 752,780   & 770,970   & 806,082   \\
(1+2) EA\_norm.            & 1,000   & 794,263   & 6\%         & 689,149   & 733,564   & 763,668   & 789,031   & 820,069   & 850,706   & 888,083   \\
(1+50) EA\_var.            & 1,000   & 824,492   & 13\%        & 670,929   & 712,694   & 752,272   & 801,097   & 868,441   & 1,009,989 & 1,052,761 \\
(1+2) EA\_var.             & 1,000   & 947,239   & 8\%         & 813,811   & 842,540   & 889,267   & 949,760   & 997,649   & 1,045,491 & 1,081,513 \\
(1+50) EA\_\textgreater{}0 & 1,000   & 829,060   & 6\%         & 721,410   & 757,243   & 799,039   & 829,887   & 863,830   & 890,942   & 914,094   \\
(1+2) EA\_\textgreater{}0  & 1,000   & 815,316   & 6\%         & 712,972   & 751,053   & 779,591   & 805,587   & 849,866   & 890,603   & 917,583   \\
RLS                        & 1,000   & 1,000,956 & 5\%         & 899,441   & 934,980   & 960,651   & 999,709   & 1,029,636 & 1,078,685 & 1,117,589 \\
\hline 
(1+50) EA\_\{r/2,2r\}      & 2,000   & 3,537,355 & 5\%         & 3,158,723 & 3,347,725 & 3,434,820 & 3,534,274 & 3,632,172 & 3,756,605 & 3,851,154 \\
(1+2) EA\_\{r/2,2r\}       & 2,000   & 4,595,036 & 5\%         & 4,169,938 & 4,274,950 & 4,401,518 & 4,608,964 & 4,731,654 & 4,901,574 & 5,071,562 \\
(1+50) EA\_half            & 2,000   & 1,875,151 & 4\%         & 1,719,176 & 1,782,887 & 1,830,853 & 1,881,673 & 1,917,520 & 1,952,149 & 2,036,373 \\
(1+2) EA\_half             & 2,000   & 2,275,140 & 4\%         & 2,076,434 & 2,151,912 & 2,204,442 & 2,276,632 & 2,347,632 & 2,397,510 & 2,434,066 \\
(1+50) EA\_norm.           & 2,000   & 1,899,807 & 5\%         & 1,719,247 & 1,777,183 & 1,847,531 & 1,902,661 & 1,948,498 & 2,007,284 & 2,062,818 \\
(1+2) EA\_norm.            & 2,000   & 2,562,111 & 4\%         & 2,280,859 & 2,407,007 & 2,485,917 & 2,569,675 & 2,629,185 & 2,683,277 & 2,753,594 \\
(1+50) EA\_var.            & 2,000   & 1,840,575 & 6\%         & 1,653,653 & 1,717,599 & 1,761,572 & 1,818,355 & 1,890,466 & 2,012,659 & 2,115,933 \\
(1+2) EA\_var.             & 2,000   & 1,951,034 & 5\%         & 1,759,004 & 1,813,671 & 1,901,088 & 1,952,594 & 1,996,883 & 2,064,192 & 2,117,979 \\
(1+50) EA\_\textgreater{}0 & 2,000   & 2,191,136 & 4\%         & 2,012,670 & 2,049,132 & 2,124,075 & 2,198,951 & 2,246,503 & 2,296,396 & 2,414,214 \\
(1+2) EA\_\textgreater{}0  & 2,000   & 2,178,207 & 5\%         & 1,982,718 & 2,033,221 & 2,111,400 & 2,176,141 & 2,235,855 & 2,305,965 & 2,368,306 \\
RLS                        & 2,000   & 1,990,912 & 4\%         & 1,840,796 & 1,887,482 & 1,934,728 & 1,988,541 & 2,041,110 & 2,074,868 & 2,157,134\\
\hline
\vspace{1ex}
\end{tabular}
\caption{Selected statistics of the fixed-target running times for the $2\,000$-dimensional \leadingones problem. AHT= average first hitting time, rsd= relative standard deviation. We have chosen to display target value $n/2=1\,000$ because this is the point after which flipping one bit becomes optimal, i.e., advantages over RLS must result from the phase before reaching this target point.}
\label{fig:LOstats}
\end{table}
\end{landscape}

\end{document}